\documentclass[runningheads]{llncs}
\usepackage{graphicx}
\usepackage{acronym}
\usepackage{amsmath}
\usepackage{amsfonts}
\usepackage{multirow}
\usepackage{array}
\usepackage{hyperref}
\usepackage{siunitx}

\acrodef{MF}[MF]{mitotic figure}
\acrodef{MC}[MC]{mitotic count}
\acrodef{WSI}[WSI]{whole slide image}
\acrodef{DL}[DL]{deep learning}
\acrodef{HE}[H\&E]{hematoxylin and eosin}
\acrodef{PHH3}[PHH3]{phosphohistone H3}
\acrodef{ROI}[ROI]{region of interest}
\acrodef{IHC}[IHC]{immunohistochemistry}
\acrodef{CCMCT}[CCMCT]{canine cutaneous mast cell tumor}
\acrodef{CMC}[CMC]{canine mammary carcinoma}
\acrodef{MIDOG}[MIDOG]{Mitosis Domain Generalization}
\acrodef{AP}[AP]{average precision}
\acrodef{ICC}[ICC]{intraclass correlation coefficient}
\acrodef{FCOS}[FCOS]{Fully Convolutional One-Stage Object Detector}
\acrodef{DI-FCOS}[DI-FCOS]{Dual-Input FCOS}

\newcolumntype{L}[1]{>{\raggedright\let\newline\\\arraybackslash\hspace{0pt}}m{#1}}
\newcolumntype{C}[1]{>{\centering\let\newline\\\arraybackslash\hspace{0pt}}m{#1}}
\newcolumntype{R}[1]{>{\raggedleft\let\newline\\\arraybackslash\hspace{0pt}}m{#1}}

\begin{document}
\title{On the Value of PHH3 for Mitotic Figure Detection on H\&E-stained Images}
%
%
\author{Jonathan Ganz\inst{1} \and
Christian Marzahl \inst{2} \and
Jonas Ammeling\inst{1} \and
Barbara Richter \inst{3} \and 
Chloé Puget \inst{4} \and
Daniela Denk \inst{5} \and
Elena A. Demeter \inst{6} \and
Flaviu A. Tabaran \inst{7} \and
Gabriel Wasinger \inst{8} \and
Karoline Lipnik \inst{3} \and
Marco Tecilla \inst{9} \and
Matthew J. Valentine \inst{10} \and
Michael J. Dark \inst{11} \and
Niklas Abele \inst{12} \and
Pompei Bolfa \inst{10} \and
Ramona Erber \inst{12,15} \and
Robert Klopfleisch \inst{4} \and
Sophie Merz \inst{13} \and
Taryn A. Donovan \inst{14} \and
Samir Jabari \inst{12} \and
Christof A. Bertram \inst{3} \and
Katharina Breininger \inst{15} \and
Marc Aubreville \inst{1}}

\authorrunning{J. Ganz et al.}

\institute{
Technische Hochschule Ingolstadt, Ingolstadt, Germany \and
Gestalt Diagnostics, Spokane, USA \and
University of Veterinary Medicine, Vienna, Austria \and
Freie Universität Berlin, Berlin, Germany \and
SeaWorld Yas Island, Abu Dhabi, UAE \and
Cornell University, Ithaca, USA \and
USAMV, Cluj-Napoca, Romania \and
Medical University of Vienna, Vienna, Austria \and
Hoffmann-La Roche, Basel, Switzerland \and
Ross University School of Veterinary Medicine, Basseterre, St. Kitts and Nevis \and
University of Florida, Gainesville, USA\and
University Hospital Erlangen, Erlangen, Germany\and
IDEXX Laboratories, Kornwestheim, Germany \and
The Schwarzman Animal Medical Center, New York, USA \and
Friedrich-Alexander-Universität Erlangen-Nürnberg, Erlangen, Germany
}
\maketitle              
\begin{abstract}
The count of mitotic figures (MFs) observed in hematoxylin and eosin (H\&E)-stained slides is an important prognostic marker as it is a measure for tumor cell proliferation. However, the identification of MFs has a known low inter-rater agreement. Deep learning algorithms can standardize this task, but they require large amounts of annotated data for training and validation. Furthermore, label noise introduced during the annotation process may impede the algorithm's performance. Unlike H\&E, the mitosis-specific antibody phospho-histone H3 (PHH3) specifically highlights MFs. Counting MFs on slides stained against PHH3 leads to higher agreement among raters and has therefore recently been used as a ground truth for the annotation of MFs in H\&E. 
However, as PHH3 facilitates the recognition of cells indistinguishable from H\&E stain alone, the use of this ground truth could potentially introduce noise into the H\&E-related dataset, impacting model performance.
This study analyzes the impact of PHH3-assisted MF annotation on inter-rater reliability and object level agreement through an extensive multi-rater experiment. We found that the annotators' object-level agreement increased when using PHH3-assisted labeling. Subsequently, MF detectors were evaluated on the resulting datasets to investigate the influence of PHH3-assisted labeling on the models' performance. Additionally, a novel dual-stain MF detector was developed to investigate the interpretation-shift of PHH3-assisted labels used in H\&E, which clearly outperformed single-stain detectors.
However, the PHH3-assisted labels did not have a positive effect on solely H\&E-based models. The high performance of our dual-input detector reveals an information mismatch between the H\&E and PHH3-stained images as the cause of this effect.
\keywords{Mitotic Figure Detection  \and Computational Pathology \and PHH3.}
\end{abstract}
\section{Introduction}
In tumor diagnosis, a crucial step is the examination of tissue samples by pathologists to derive important information related to the tumor and its appropriate treatment. One factor of interest is the proliferation fraction of the respective tumor, which can be assessed through the number of cells undergoing cell division, depicted by \acp{MF} \cite{Baak2005}. 
The \ac{MC}, defined as the number of \acp{MF} within a standardized area of ten consecutive high-power fields \cite{Meuten2016}, is part of various tumor grading systems in human \cite{ELSTON1991,Louis2021} as well as veterinary medicine \cite{kiupel2011proposal,pena2013prognostic}. However, counting \acp{MF} is a task known to have a low inter-rater agreement \cite{Veta2016,bertram2022computer}.
The use of slide scanners has enabled the digitization of entire slides into \acp{WSI}, which can be evaluated automatically using \ac{DL} methods. In the context of \ac{MF} detection, \ac{DL} algorithms already demonstrated human-like performance \cite{bertram2022computer}. Nevertheless, those algorithms rely on large amounts of annotated data, and the annotation quality is known to affect the performance of the trained \ac{DL} model \cite{Bertram2020,Wilm2021}.
In contrast to the \ac{HE} stain, which is the standard stain used in histopathology, the mitosis-specific antibody \ac{PHH3} specifically highlights the cell nucleus during mitosis \cite{Duregon2015}. Staining against PHH3 is a validated method with diagnostic and prognostic significance, emphasizing its utility in the assessment of the MC in various tumor types \cite{Colman2006,Duregon2015,Voss2014,Laflamme2020,Skaland2009}.
Furthermore, different studies reported that \ac{MC} derived solely by \ac{PHH3} leads to lower variability of the \ac{MC} among different raters compared with the \ac{MC} acquired through \ac{HE} \cite{Laflamme2020,Duregon2015}. Nevertheless, \ac{HE} is the routine stain in pathology, and thus detectors intended for clinical use have to be applicable to this stain. 
In order to still be able to leverage the advantages of \ac{PHH3} staining to improve the quality of \ac{HE}-stained \ac{MF} datasets, staining against \ac{PHH3} has been used as an assistance for the annotation of \acp{MF} in \ac{HE} by de-staining \ac{HE}-stained slides and re-staining them against \ac{PHH3} \cite{tellez2018whole,bertram2022computer}. The re-stained slides are then digitized and registered with their \ac{HE} counterparts. However, \ac{PHH3} is more sensitive to early mitotic phases and less sensitive to late phases of the cell cycle \cite{Voss2014}. As a result, the \acp{MF} highlighted in slides stained against \ac{PHH3} differ from those recognizable in \ac{HE}, leading to an elevated (though biologically more accurate) \ac{MC} \cite{Duregon2015}. 
Our hypothesis is that providing an annotator with \ac{PHH3} assistance, as described above, may introduce a potential bias that results in the inclusion of \acp{MF} in the dataset that would not have been annotated with only \ac{HE} available. This, in turn, may result in the inclusion of \acp{MF} that cannot be identified with only the \ac{HE}-stained slides available, leading to an information mismatch and, from the perspective of the \ac{HE}, noisy \ac{MF} labels.
Our main contributions are:

\begin{itemize}
    \item Extensive multi-rater experiments were conducted to show \ac{PHH3}-assisted annotation's superior labeling consistency and to generate high-quality data\-sets to evaluate our methods.
    \item Using our novel dual-stain MF detector, we show the principal superiority of PHH3-assisted labels, given that both stains are available.
    \item We show that PHH3-assisted labels can cause interpretation shifts in H\&E when training MF detectors. This implies that a 'biologically accurate' PHH3-assisted ground truth may not improve performance if information mismatch is ignored.
\end{itemize}

\section{Datasets}
Two different datasets were used in this study, one for the training of the object detectors and one for the multi-rater experiment. Both datasets utilize corresponding \ac{ROI} pairs of tumor tissue stained with two stains: The source slides were initially stained with \ac{HE}, digitized, and subsequently de-stained and re-stained with \ac{PHH3} before being digitized again. The resulting \acp{WSI} were then registered using a registration algorithm for \acp{WSI} \cite{marzahl2021}. From the original \acp{WSI}, the \acp{ROI} were selected by two pathologists 
based on tissue and scan quality and the area of perceived highest mitotic activity.

\subsection{Image Dataset for the Annotation Study}
The dataset used in the annotation study consists of 20 \acp{ROI} representing four different tumor types, two of which were of human origin and of two of veterinary origin. Tumors of different tumor types and species were included in order to draw broader conclusions from the results of the study. Five samples each of human astrocytoma and meningioma were collected from the diagnostic archive of \textit{anonymized hospital}, after prior ethics approval (No. ANON). The slides were digitized using an Hamamatsu NanoZoomer S60 at 40$\times$ magnification. From the diagnostic archive of \textit{anonymized university} five samples each of \ac{CCMCT} and of \ac{CMC} were collected. For these slides, no ethics approval was needed. These slides were digitized with a 3DHistech Pannoramic Scan II at 40$\times$ magnification. Each \ac{ROI} was selected to cover $2.37\,\si{mm}^2$ of tissue, which is equivalent to approximately ten high power fields \cite{Meuten2016}. In the remainder of this paper we refer to the dataset generated by annotating these images as \textbf{study dataset}.

\subsection{Development Dataset}
To examine the impact of \ac{PHH3}-assisted \ac{MF} annotation on the performance of \ac{MF} detectors, we trained detectors with annotations acquired only using \ac{HE}-stained slides and with annotations that were generated through \ac{PHH3}-assisted annotation. 
For this purpose, we used a datatset consisiting of ROIs of multiple tumors and species from different laboratories. The dataset contained ten samples of \ac{CCMCT}, nine samples of \ac{CMC}, ten sampes of canine hemangiosarcoma, nine samples of feline lymphoma, ten samples of feline soft tissue sarcoma, three samples of human astrocytoma, ten samples of human bladder cancer, nine samples of human colon carcinoma, ten samples of human melanoma, and four samples of human meningioma. We refer to this as the \textbf{development dataset}.
Two different ground truth definitions were available for this dataset. The first definition relied solely on \ac{HE}-stained slides, while the second was created with \ac{PHH3}-assisted labeling. 
The \ac{HE}-only annotations were created as described in~\cite{Bertram2020} by three pathologist with at least five years of experience in \ac{MF} identification assisted by a \ac{DL} model with high recall that screened the slides for \ac{MF} candidates. To be accepted as a \ac{MF}, at least two pathologists had to agree upon a candidate.
The \ac{PHH3}-assisted annotations were created by a single expert using an open source web-based annotation server \cite{marzahl2021exact} where the two corresponding stains could be superimposed on each other with variable transparency. This way, the expert was able to assess the \ac{IHC} label present in the \ac{PHH3}-stained slide and the morphological features visible in the \ac{HE} slide at once. If the registration was not perfect, e.g., due to tissue deformation, the position of the respective cell in the \ac{HE}-stained slide was annotated. Cells that had a positive \ac{IHC} label but that were lacking \ac{MF} morphology in the \ac{HE} stain were not annotated, as these cells are not identifiable as \ac{MF} in the \ac{HE}-stained slides.

\section{Methods}
We investigated the value of \ac{PHH3}-assisted \ac{MF} annotation from two different perspectives. The impact on the inter-rater agreement was investigated via a multi-rater experiment. In a second experiment, the datasets resulting from this experiment were used to evaluate \ac{MF} detectors trained on labels derived with and without \ac{PHH3} assistance.

\subsection{Human Expert Mitotic Figure Annotation Study}

To investigate the impact of co-registered \ac{PHH3} slides on inter-rater agreement, a study was conducted with 13 pathology experts. The study consisted of two phases, with a four week washout period to prevent participants from recalling the cases. To further prevent bias, the slides were presented in randomized order and under different names in each phase. During the initial phase, participants only had access to \ac{HE}-stained \acp{ROI}.
In the subsequent phase, participants could overlay co-registered \ac{PHH3}-stained \acp{ROI} on the \ac{HE}-stained image with adjustable transparency, allowing for simultaneous examination of both stains. Additionally, participants were given three different labels to choose from when annotating \acp{MF} in the second phase, depending on whether a \ac{MF} was identifiable in both \ac{HE} and \ac{PHH3}, only in \ac{HE}, or only in \ac{PHH3}. In this paper, only annotations of the first two classes are considered, as these match the annotation semantics of the \ac{PHH3}-assisted ground truth of the MIDOG dataset. For imperfect registration, participants were instructed to annotate the position of the respective cell in the \ac{HE}-stained \ac{ROI}. Before the second phase a pathologist highly experienced in \ac{PHH3}-assisted labeling conducted an online training with the participants.

\subsection{Detection Architectures}
\begin{figure}[ht]
    \centering
    \includegraphics[width=1\linewidth]{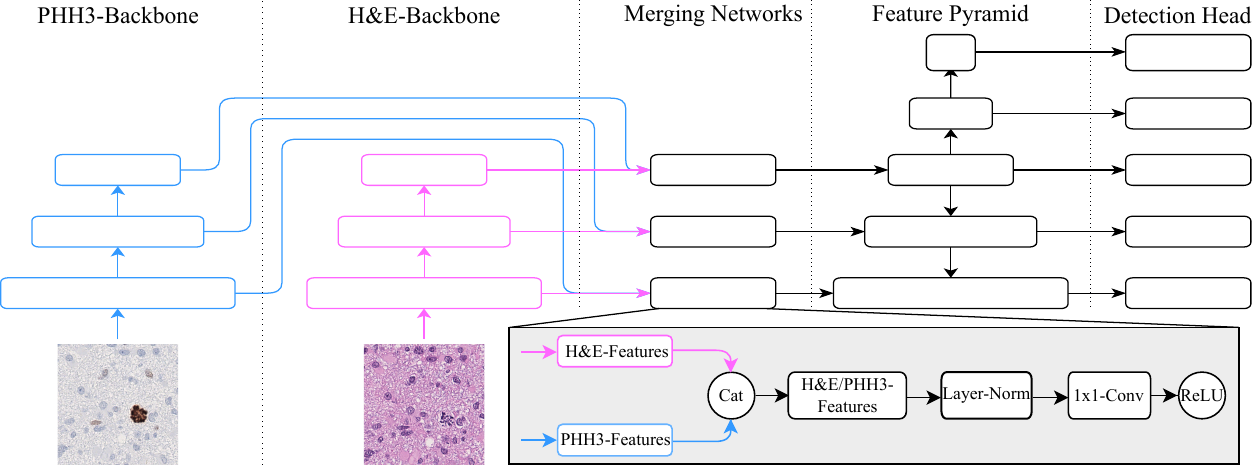}
    \caption{Architectural overview of our dual-stain \ac{MF} detector.}
    \label{fig:architecture}
\end{figure}

\noindent To assess the impact of \ac{PHH3}-assisted labeling on the performance of object detectors, we used the \ac{FCOS} \cite{tian2019fcos}. Our hypothesis was that the annotation assisted by \ac{PHH3} will introduce \acp{MF} into the dataset that are not distinguishable in \ac{HE} alone. The solely \ac{HE}-based detector lacks the information from the \ac{PHH3}-stained section, which causes an interpretation shift in the labels and introduces noise into the \ac{HE} dataset. This noise could potentially impede the performance of models trained on such a dataset. If this is correct, a detector that is able to use the information from both stains should show better results on such a dataset than a pure \ac{HE} detector.
To test the hypothesis we extended the \ac{FCOS} detector into a \ac{DI-FCOS} detector as depicted in Fig. \ref{fig:architecture}, which was trained on the corresponding \ac{HE} and \ac{PHH3} patches simultaneously. The \ac{DI-FCOS} model consists of two ResNet18 backbones. Feature fusion is achieved via mid-fusion, i.e., information from different input modalities is combined at an intermediate stage within the network \cite{qingyun2022cross}. In particular, the features of each ResNet level are fused before they are forwarded to the feature pyramid, using one merging network for each input level of the feature pyramid. Let $\mathbf{H}\in\mathbb{R}^{C \times H \times W}$ and $\mathbf{P}\in\mathbb{R} ^{C \times H \times W}$ be the feature maps of a level of the \ac{HE} and \ac{PHH3} backbone, where $C$ represents the number of channels of the respective level and $H$ and $W$ are the sizes of the feature maps which depend on the size of the input image. Then the fused features $\mathbf{F}\in\mathbb{R}^{C \times H \times W}$ are computed by $\mathbf{F} = \text{{ReLU}}(\text{{Conv}}(\text{{LayerNorm}}(\text{{Cat}}(\mathbf{H}, \mathbf{P}))))$, where $\text{Cat}$ denotes a concatenation of the feature vectors along the channel dimension and Conv is a $1 \times 1$ convolution which halves the number of input channels to $C$ after the concatenation. The rest of the network follows the standard \ac{FCOS} architecture as described in \cite{tian2019fcos}. To confirm that a difference in model performance between the \ac{FCOS} and \ac{DI-FCOS} models is not due to the \ac{DI-FCOS} models' higher number of parameters, we also compare it to an \ac{FCOS} detector with a ResNet101 backbone. For all experiments, the feature maps from the second, third, and fourth blocks of the respective backbone were used to construct the feature pyramids for both the standard \ac{FCOS} and the \ac{DI-FCOS} model. We used a fixed learning rate of $10^{-4}$ and AdamW as the optimizer. All models were trained until convergence, which was observed using the \ac{AP} metric on the validation set, which we also used for early stopping and for model selection. All models were trained on patches with a height and width of 512 pixels, and patches were selected so that at least $50~\%$ of the training patches contained \acp{MF}. A standard image augmentation pipeline was used during the training of each object detector in this study. 

\section{Results}
We first describe the results of the human-rater experiment as the resulting dataset was part of the evaluation of the object detectors. 
\begin{figure}[h]
    \centering
    \includegraphics[width=1\linewidth]{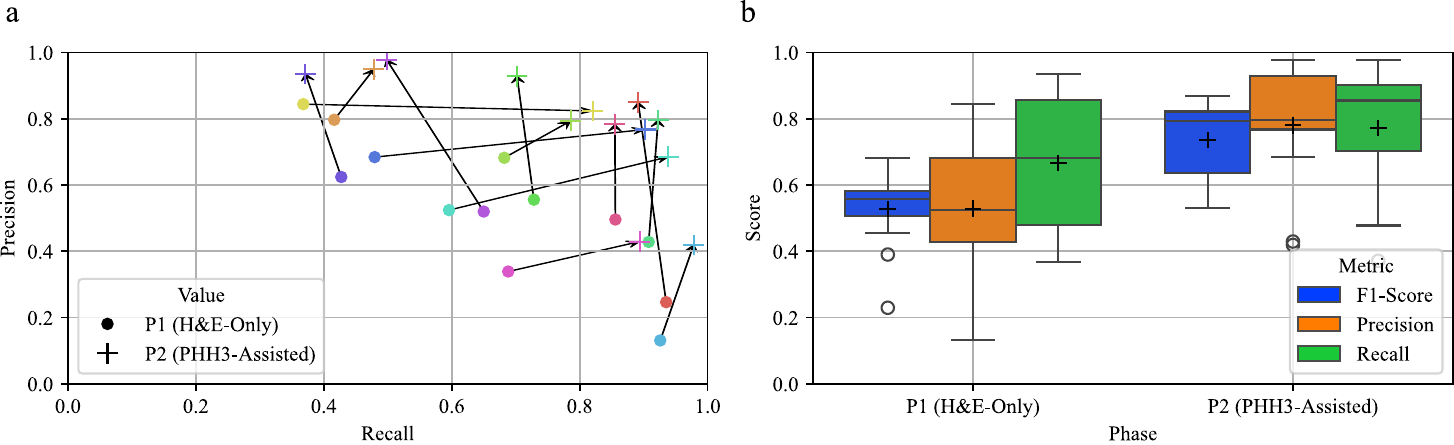}
    \caption{Figure (a) displays the precision and recall of each rater plotted against the consensus of the remaining raters of phase one and three. The results of the first (P1) and second (P2) phases of the study are marked by a dot and a cross, respectively. Each rater is represented by a different color. The F1 value, precision, and recall of each rater against the consensus of the remaining raters is given in Figure (b). Means are indicated by the black crosses.}
    \label{fig:res_human_rater_experiment}
\end{figure}
To measure agreement at the object level, we compared each rater's annotation on images of the \textbf{study dataset} to the consensus of all other raters' annotations for each phase of the experiment. We formed the consensus by matching the annotations of the remaining raters for each image using a distance-based clustering approach. An annotation was added to a cluster if it was no more than 7.5 micrometers away from the center of the cluster, approximately corresponding to the diameter of a nucleus \cite{aubreville2023mitosis}. 
A cluster was considered a MF if it contained annotations from at least six raters. Using a higher threshold could result in suboptimal results because raters tend to miss \acp{MF} that are difficult to identify \cite{Veta2016,Bertram2020}. Hence, a higher threshold may lead to the exclusion of \acp{MF} that are likely to be overlooked. For the creation of the \ac{HE}-only label set of the development dataset, a \ac{DL} model with high recall was used to overcome this problem. Although this threshold is a hyperparameter that can influence the results, the same trend was observed across different thresholds (see supplementary Figure S1). The agreement between the raters' annotations and the described consensus was then measured via Dice Similarity Coefficient/F1-score, as proposed in \cite{Veta2016}, precision, and recall. Additionally, we evaluated the inter-rater reliability of the MC using the \ac{ICC}.
Figure \ref{fig:res_human_rater_experiment}~(a) shows the precision and recall values for each rater in the different phases of the study. Each rater demonstrated an increase in either recall, precision, or both when comparing the results of phase P1 to P2. If there was a slight decrease in either precision or recall in phase P2, it was always accompanied by a substantial increase in the other metric. The results over all raters are given in Figure \ref{fig:res_human_rater_experiment}~(b). Each metric increased by a large margin from P1 to P2. In particular we found the average F1-score to increase from $0.53 \pm 0.11$ to $0.74 \pm 0.11$, the average precision from $0.53 \pm 0.20$ to $0.78 \pm 0.17$, and the average recall from $0.67 \pm 0.19$ to $0.77 \pm 0.19$. Furthermore, the \ac{ICC} increased from $0.90$ in P1 to $0.99$ in P2.\\
Second, we investigated the performance of \ac{MF} detectors that were trained with \ac{HE}-only and with \ac{PHH3}-assisted labels. To increase the statistical informativeness of the results we trained each object detector in a five-fold Monte Carlo cross-validation scheme. For this, the \textbf{development dataset} was randomly split into $70 \%$ training, and $15 \%$ validation and test cases five times. To ensure comparability of the results, all models were trained and tested on the same five splits. Furthermore, we evaluated the resulting models on the study datasets derived from P1 (\textbf{\ac{HE}-only}) and P2 (\textbf{\ac{PHH3}-assisted}) of the human rater experiment. The results of both evaluations are given in Table \ref{tab1}.  The \ac{AP} was used to measure the object detection performance.
\begin{table}[h]
\caption{Results of the \ac{FCOS} and \ac{DI-FCOS} models on the test sets of the five fold cross-validation on the different label sets of the development dataset, and results of the inference of the models on the study dataset. Given are the means and standard deviations of the average precision as a result of cross-validation. We found only a minor impact of the threshold of annotators in the ground truth, see Figure S3.}
\resizebox{\linewidth}{!}{
\begin{tabular}{ccc|C{2.5cm}|C{2.5cm}|C{2.5cm}|C{2.5cm}}
\hline
Test dataset                 & Model            & Parameters\, & \multicolumn{2}{c|}{\,Trained using HE-only labels \,} & \multicolumn{2}{c}{\,Trained using PHH3-assisted labels\,} \\ \hline
& & &\multicolumn{2}{c|}{Evaluation labels} &\multicolumn{2}{c}{Evaluation labels }      \\
&                  &                & HE-only& PHH3-assisted&HE-only& PHH3-assisted\\ \hline
\multirow{3}{*}{\begin{tabular}[c]{@{}c@{}}Development\\ dataset\\\end{tabular}} & \ac{FCOS} (ResNet18)  & 19.0\,M         &$0.70\pm0.03$& $0.64\pm0.02$& $0.71\pm0.03$& $0.68\pm0.04$\\
                             & \ac{FCOS} (ResNet101) & 51.0\,M         &$\mathbf{0.74}\pm0.04$& $0.68\pm0.05$& $0.71\pm0.04$& $0.69\pm0.06$\\
                             & \ac{DI-FCOS} (ResNet18)         & 39.9\,M         &$\mathbf{0.74}\pm0.04$& $0.73\pm0.05$& $0.72\pm0.04$& $\mathbf{0.79}\pm0.05$\\ \hline
\multirow{3}{*}{\begin{tabular}[c]{@{}c@{}}Study\\ dataset\end{tabular}}       & \ac{FCOS} (ResNet18)  & 19.0\,M         &$0.64\pm0.02$& $0.58\pm0.03$& $0.61\pm0.04$& $0.61\pm0.02$\\
                             & \ac{FCOS} (ResNet101) & 51.0\,M         &$0.66\pm0.03$& $0.60\pm0.02$& $0.62\pm0.03$& $0.60\pm0.04$\\
                             & \ac{DI-FCOS} (ResNet18)          & 39.9\,M         &$0.66\pm0.04$& $\mathbf{0.72}\pm0.06$& $0.61\pm0.05$& $\mathbf{0.81}\pm0.05$                       
\\\hline
\end{tabular}}
\label{tab1}
\end{table}
We found that the single input \ac{FCOS} models performed better on the \ac{HE}-only labels compared to the \ac{PHH3}-assisted labels, regardless of whether they were trained with \ac{HE}-only labels or with \ac{PHH3}-assisted labels. 
The \ac{DI-FCOS} models outperformed the \ac{FCOS} models on the PHH3-assisted label sets of both datasets. On the HE-only label sets of both datasets, the \ac{FCOS} model with the larger backbone performed on-par with the \ac{DI-FCOS} model. Overall, the highest performance was observed when the \ac{DI-FCOS} models were trained and tested on \ac{PHH3}-assisted labels, with a mean AP of $0.79 \pm 0.05$ on the development dataset and a mean AP of $0.81 \pm 0.05$ on the study dataset (see Figure S2 for examples). The performance of these models was much lower when tested on the \ac{HE}-only labels. 

\section{Discussion}
This study demonstrates that the inter-rater reliability and agreement of \ac{MF} annotations substantially improve with the use of co-registered \ac{PHH3}-stains of the exact same slide as an annotation assistance. However, the performance of the single-input object detectors on the \ac{PHH3}-assisted labels also confirmed our hypothesis that the \ac{PHH3} assistance causes \acp{MF} to be included in the data set that are not distinguishable in the \ac{HE} alone. Given the only marginal difference between both single input models when tested on the \ac{HE}-only labels, it appears that including those \acp{MF} for model training has only a minor effect. However, if a \ac{PHH3}-assisted label set is used to evaluate an \ac{MF} detector, this may underestimate the performance of the detector. This is likely due to a low recall on cells that are non-identifiable as \acp{MF} in \ac{HE}. Since the \ac{IHC} label is dependent on the biological process of cell division, the \ac{PHH3}-assisted ground truth can be considered a more accurate estimate of the actual biological ground truth. Therefore, while a \ac{MF} detector trained with \ac{PHH3}-assisted data may perform less favorably on a \ac{HE}-only ground truth, its predictions could be closer to the biological ground truth. The results of our \ac{DI-FCOS} detectors on the \ac{PHH3}-assisted labels demonstrate that high performance on a \ac{PHH3}-assisted label set is possible when information from both modalities is available. 
Although \acp{MF} in the \ac{PHH3}- and \ac{HE}-stained sections may not always align perfectly (see Figure S2), the \ac{DI-FCOS} results demonstrate robustness against this displacement. Fine registration of patches before inputting them into the detector could potentially enhance the results even further.
Considering that similar results were achieved with both label sets in the development dataset, it is reasonable to conclude that \ac{PHH3}-assisted labeling with only one rater could replace a complex multi-rater approach as described in \cite{Bertram2020}. Hence, future research can investigate how to use its potential for labeling in \ac{HE} without any negative effects.


%
%
%
%
\bibliographystyle{splncs04}
\bibliography{my_bib}
\end{document}


\section{Supplementary Materials}
\begin{figure}
    \centering
    \includegraphics[width=0.8\linewidth]{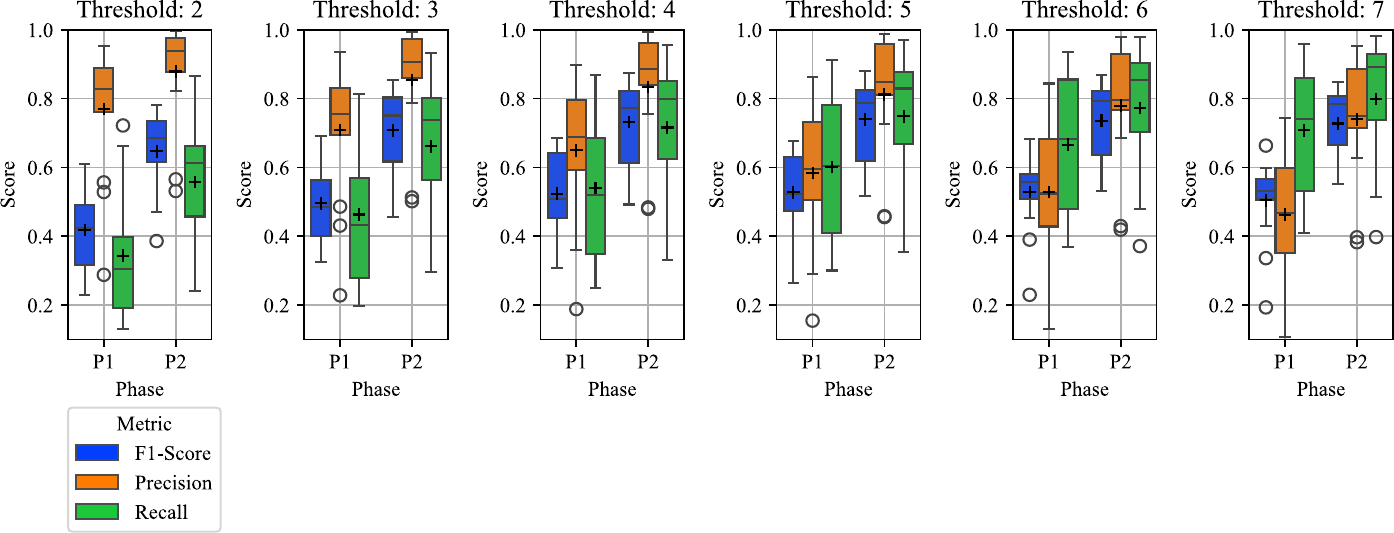}
    \caption{Comparison of F1-score, precision, and recall for individual raters against the multi-rater consensus. The analysis considers varying thresholds for the number of raters required to agree on a \ac{MF} in forming a consensus. Higher inter-rater agreement is found for PHH3-assisted annotation (P2) compared to HE-based annotation (P1) for different consensus thresholds.}
    \label{fig:supp1}
\end{figure}

\begin{figure}
    \centering
    \includegraphics[width=0.8\linewidth]{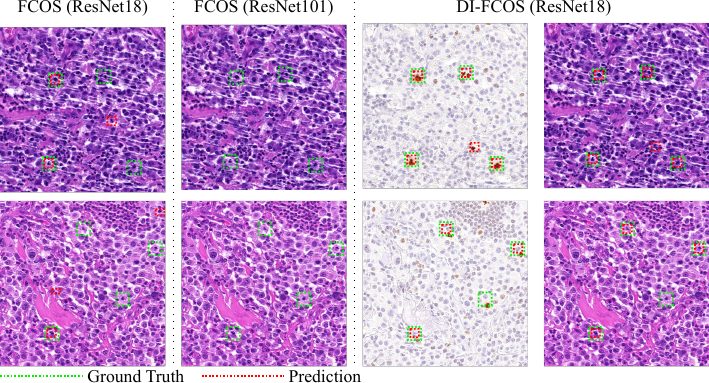}
    \caption{Qualitative detection examples of the FCOS and DI-FCOS models on the PHH3-assisted labels of the sutdy dataset: Despite imperfect alignment of HE- and PHH3-stained patches (lower row), the DI-FCOS model successfully detected MFs.}
    \label{fig:enter-label}
\end{figure}

\begin{figure}
    \centering
    \includegraphics[width=0.8\linewidth]{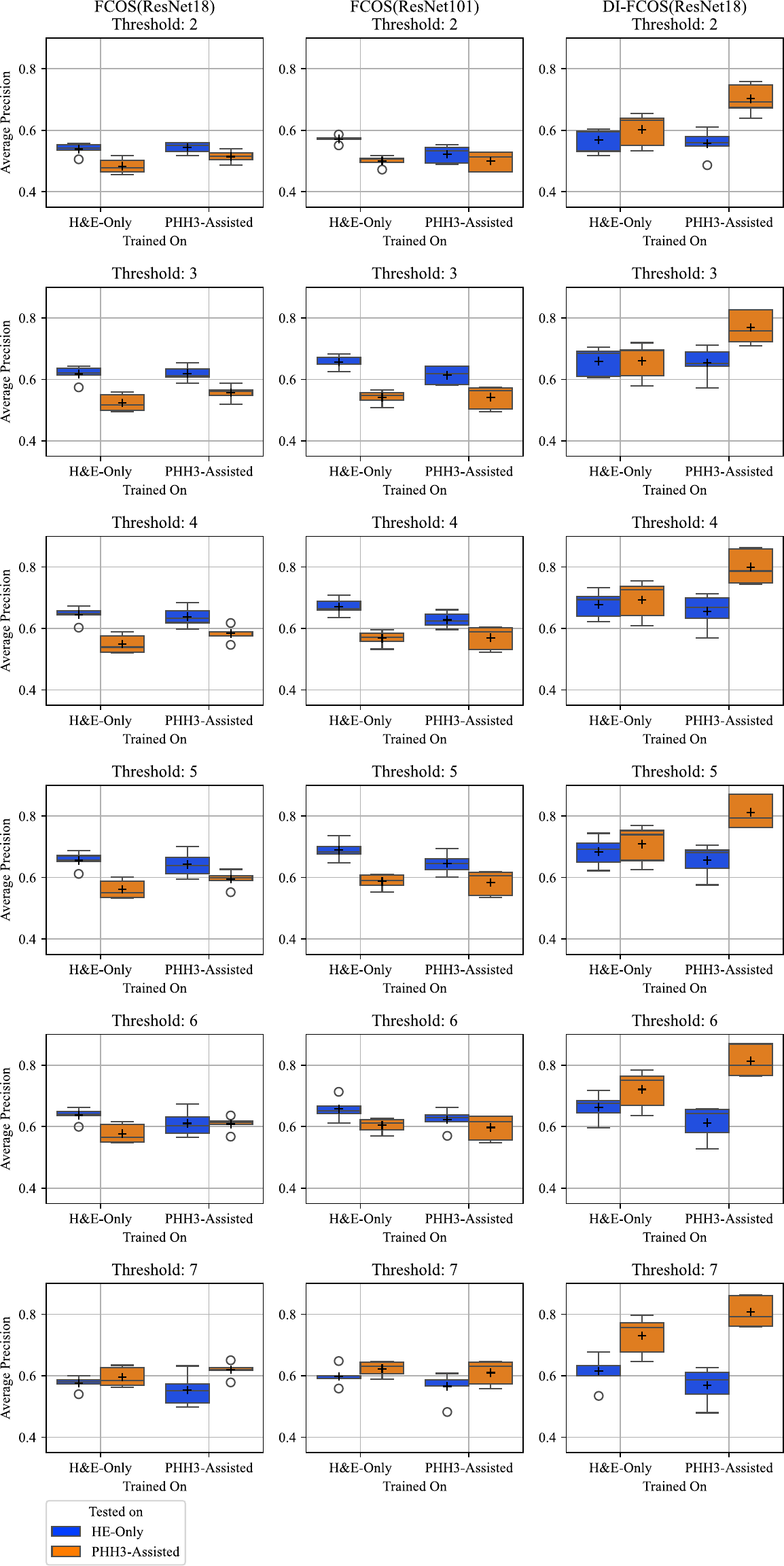}
    \caption{Results of five-fold cross-validation of the FCOS and DI-FCOS models on the study dataset, as a function of the number of raters necessary to agree upon an object to be counted as a ground truth (GT) MF from two (top) to seven (bottom). For higher thresholds, the performance of the detector drops, as the number of GT MFs drops due to a high omission rate in the H\&E annotations, while the detector performance is not altered significantly when using the PHH3-assisted annotation, as the omission rate is strongly reduced and rater agreement increased.}
    \label{fig:supp2}
\end{figure}